# Detecting Botnets Through Log Correlation

Yousof Al-Hammadi and Uwe Aickelin
School of Computer Science University
of Nottingham, Nottingham, NG8 1BB,
UK
yxa,uxa@cs.nott.ac.uk

*Abstract*— Botnets, which consist of thousands of compromised machines, can cause significant threats to other systems by launching Distributed Denial of Service *(DDoS)* attacks, keylogging, and backdoors. In response to these threats, new effective techniques are needed to detect the presence of botnets. In this paper, we have used an interception technique to monitor Windows Application Programming Interface *(API)* functions calls made by communication applications and store these calls with their arguments in log files. Our algorithm detects botnets based on monitoring abnormal activity by correlating the changes in log file sizes from different hosts.

*Keywords*-IRC; DDoS; Bots; Botnets; API function calls

## I. INTRODUCTION

Recently, an explosive growth of coordinated attacks has been noticed [1][6]. This kind of attack is performed by using Internet Relay Chat (*IRC*) networks to control compromised machines (*zombies*) and establish a distributed attack against other systems. These zombies are infected by a piece of malicious code named a *bot* [1][6]. Malicious bots are programmed to respond to various instructions generated by the attacker.

A collection of compromised machines that are connected to a single channel on *IRC* networks forms a *(Botnet)*. These machines can be controlled remotely by the attacker via command and control *(C&C)* to perform malicious activities such as *DDoS* attack. A *DDoS* attack is established when many bots start to flood other networks by sending them large numbers of packets. Current botnets usage trends focus on email spamming/bombing, steal system information, program termination and extorting money from on-line businesses [3][6][7].

Most current bots are implemented to use a centralized network, which allow them to receive instructions from a central point. This makes the process of tracing the bot herder (i.e. the attacker) a relatively easy task. A more dangerous threat appears when the bot herder designs his/her bots to work in a Peer-to-Peer (*P2P*) environment, which makes the tracing process more complex. We focus on detecting botnets that use a centralized network. Detecting botnets in a *P2P* network forms our future work.

There have been several studies in detecting and tracking botnets using a non-productive resource *(honeypot)* and analysing traffic patterns [2][3]. A honeypot is a system resource that is not meant to provide any services to legitimate users. One problem with using honeypots is that they cannot detect suspicious traffic without receiving activity directed against them [11]. In addition, monitoring *IRC* traffic patterns on standard ports used by *IRC* clients generates some false negatives since bots can run on non-standard ports. Moreover, there are no simple characteristics of communication channels that can be used for detection. For instance, the outgoing connections have different lengths and the number of bytes transferred per connection is not fixed [2].

To address these problems, our aim is to detect botnets by monitoring the change of behaviour in log file sizes across several hosts and find the correlation between these changes. This is due to the fact that bots are responding to the commands simultaneously which produce the same rate of change in each log file. Our approach does not process *IRC* traffic searching for specific patterns. Therefore, the amount of processing time required to detect botnets will be reduced. In addition, we do not monitor standard ports and worry about encrypted traffic, because our approach monitors the change of behaviour in the system not the content of each packet.

We discuss the mechanism of collecting traffic from hosts in section two. Section three explain how to design and implement such system. We present our results in section four as well as explain our idea of detecting botnets. Finally, we will conclude and discuss our future work in section five.

## II. DATA COLLECTION

Our main goal is to detect botnets by monitoring the change of behaviour of log file sizes from different hosts and find the correlations between these log file sizes. To achieve this goal, we use a technique implemented by [5] to intercept *API* socket function calls produced by communication applications to generate our data. The intercepted *API* socket function calls and their arguments are stored in log files.

We use a system-wide intercepting technique [8] which monitors all threads currently running on the system to intercept *API* socket function calls such as *send(), sendto(), recv(), recvfrom(), or connect()*. One way to intercept an *API* socket function calls is to implement a Dynamic Link Library (*DLL*) file which replaces the target function to be intercepted (e.g. recv()) with an intercepted function (e.g. myrecv()) and then inject the *DLL* file into the address space of target process [4][10] (e.g Internet Relay Chat client - mIRC).

Once the *DLL* is loaded into the target process (mIRC), it modifies the address of the target function (e.g. recv()) in the target process (mIRC) so that it jumps to the replacement function in the DLL (myrecv()).

**Algorithm 1**: Correlation Algorithm

```
forall logfiles do
    read file sizes of each logfiles
    if all current file sizes did not change from the previous
    sizes then
        outfile = generate zeros  correlation
    else if all current file sizes changed from the previous sizes
    then
        outfile = generate ones  correlation
    else
        /* some current file sizes changed */
        outfile = generate uncorrelation
    end
while !eof.outfile do
    if zeros correlation || ones correlation then
        CV ++        /* Correlated Value (CV) */
    else             /* Uncorrelated Value (UCV) */
        UCV ++
    end
end
if CV > Threshold then
    suspicious activity is detected
end
```

## III. DESIGN AND IMPLEMENTATION

System-wide interception can be used to monitor communication applications. For example, it can be used to intercept *API* socket functions. Using this, we describe our algorithm in more details in this section.

First, we intercept *API* socket function calls used by communication programs, and store them with their arguments into a log file. During this, another program is used to record the change of log file size. This record is made every second for a period of time $t$. We assume that the log files are protected and the attacker can not erase the log files. After a time $t$, the recorded data is passed to the analyser. The analyser reads the recorded data for each host and checks to see if there is a change from current state (e.g. $t_2$) with previous state (e.g. $t_1$) for all recorded data from different hosts. If there is a change, a value of one is produced, otherwise, a value of zero is produced. Note that we are not considering the amount of change at the moment. Both all zeros (i.e. no change between log files is made from different hosts) and all ones (i.e. all log files from different hosts are changed) mean correlation between data.

For example, if there is no change between data sets at time $t_1$, (logfile1=0,logfile2=0,logfile3=0,...), then we have *zeros* change correlation. If there are changes in all data sets at time $t_1$, (logfile1=1,logfile2=1,logfile3=1,...), we have *ones* change correlation. Otherwise, an uncorrelated event is recorded. We will consider the amount of change between data sets in our future work. Our correlation algorithm is shown in Algorithm 1.

### A. Full details of Architecture

To perform our experiments, we set up a small virtual *IRC* network on a VMWare machine. The VMWare machine runs under a Windows XP P4 SP2 with a 2.4GHz processor and 1GB RAM. The virtual *IRC* network consists of four machines. One machine run Windows XP Pro SP2 and it is used as an *IRC* server. The remaining machines run Windows XP Pro SP2 and have *IRC* clients. Different experiments are conducted to analyse normal behaviour and abnormal behaviour. Each experiment was running for 10 minutes in order to collect a reasonable amount of traffic.

### B. Experiments

We conducted some initial experiments to determine if network statistics Logs alone are sufficient to detect bots. For example, we monitor the change of behaviour of Internet Explorer *(IE)* vs. *sdbot* [9]. The results show that there is a sudden increase in log file size when the bot herder uses his bot to perform UDP, or ICMP flood against other systems. On the other hand, *IE*, which is used for browsing, checking emails, and other services not including downloading/uploading files, shows a smooth increase in log file size. After that, we investigate the normal behaviour of an mIRC clients vs. the *sdbot*. Monitoring changes of behaviour of normal mIRC clients and *sdbot* shows that there is a sudden change in the case of transferring large files between mIRC clients similar to bot attack. In order to distinguish normal behaviour of mIRC clients and abnormal behaviour of bots, we analysed two cases: the normal case and the attack case. In the normal case, we analysed two scenarios:

- Three users having normal conversation.
- Three users having normal conversation and sending files to each other.

In the attack case, we analysed two scenarios:

- Three bots join an *IRC* channel and remain idle for two minutes. After the idle period, the bots start to receive commands from their master (not including flood attack commands).
- Three bots join an *IRC* channel and remain idle for two minutes. After the idle period, the bots start to receive commands from their herder including flood attack commands.

The generated results are passed to our correlation algorithm to distinguish normal behaviour and abnormal behaviour. Note that we have normalised the x-axis to 100 bytes in order to make the graphs more comparable. The next section explain our results in more detail.

## IV. RESULTS

We monitor the change of behaviour between mIRC clients and *sdbot*. The results in Figure 1 show that it might be difficult to distinguish the normal behaviour from malicious behaviour because there is a noticeable change of log file size generated during a file transfer. We also notice that it is not sufficient to just look at network statistics. Therefore, we use our correlation algorithm to distinguish between normal and abnormal behaviour.

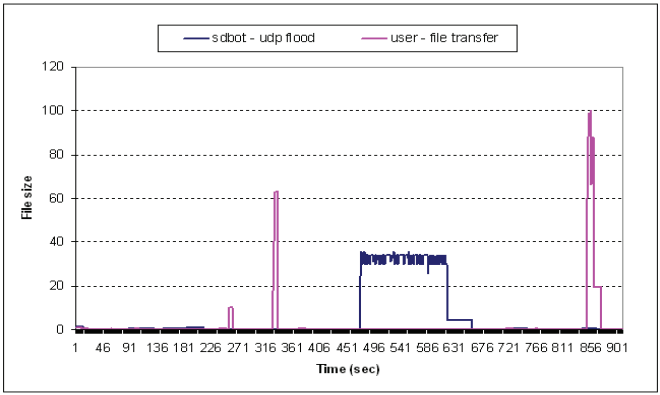

Fig. 1. Change of log file size (a user transfers files vs. a bot using UDP and ICMP flood. (100 ≡ 275085 bytes)

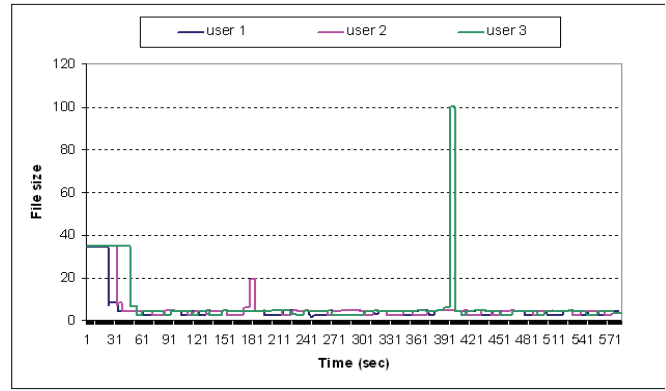

Fig. 3. Normal users behaviour with sending files. (100 ≡ 7248 bytes)

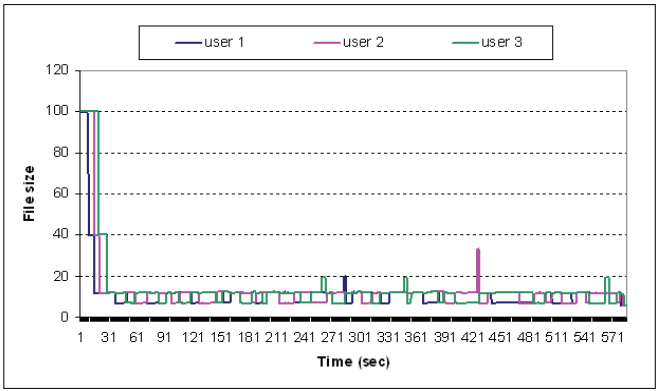

Fig. 2. Normal users behaviour without sending files. (100 ≡ 2754 bytes)

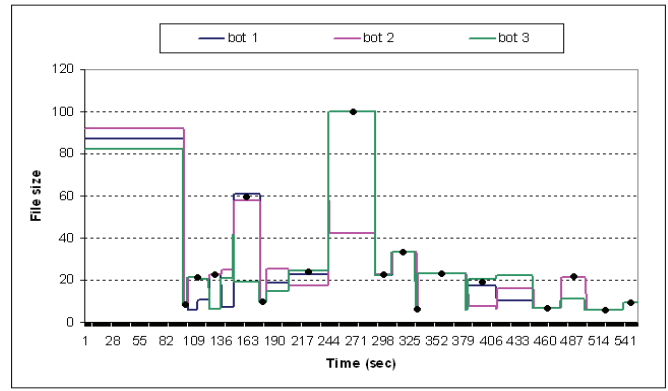

Fig. 4. Attack behaviour without flood. (100 ≡ 4121 bytes). The dot points represents a high correlation between bots

*A. Botnet Detection through Distributed Log Correlation*

The results from the previous experiment show that sometimes it is difficult to distinguish the normal behaviour from malicious behaviour, e.g. when there is a sudden change of log file size. Therefore, we present our correlation detection scheme to distinguish between these two cases.

The basic idea is to find correlated events in different hosts. Since we are dealing with botnets, there is a high probability of having correlated events such as sending similar amounts of data to a bot herder that occur within a specified time, or generating similar amounts of traffic to attack other systems. As a result, a high correlation between events is generated. A high correlation represents malicious activity, while a low correlation represents normal activity.

We investigate the normal scenario of three users having normal conversation and using some IRC commands without transferring files to each other. The results show that there is a low correlation generated from the three users (Figure 2). We also investigate the normal scenario of transferring files between users. The results show that even with a sudden change in log file size generated due to file transfer by user 3, we still notice a low correlation between data (Figure 3).

After simulating the normal cases, the three machines were infected by *sdbot05b*. This represents the two attack cases.

In the first experiment, we investigate the attack scenario of three bots receiving commands from their bot herder. No flood attack commands were received. We notice that the generated data is small but there is a high correlation between the changes of log file sizes (Figure 4). In the second attack scenario, the bots receive flood commands from their herder. The results show that there is an obvious malicious activity in the network. This can be seen from the sudden change of the amount of data generated and the high correlation between the changes of log file sizes (Figure 5).

The results from the correlation algorithm is shown in Figure 6. The x-axis represents the normalized data while the y-axis represents the conducted experiments. We can see from the figure that we have a large number of uncorrelated events in the normal case. This represents a normal behaviour in our case since users are responding randomly to others. On the other hand, the uncorrelated events in the attack case are generated due to the fact that sometimes there is a delay of responding to the bot herder's commands. We also notice that there is a large number of correlated events in the normal case. There are many reasons for this. The first reason is that we are running our experiments in virtual machines and switching between virtual machines takes some time. Another reason is that we are recording our data every second. Since, we have only one person (simulating to be three), recording data every second produces a large number of correlated events in the normal case.

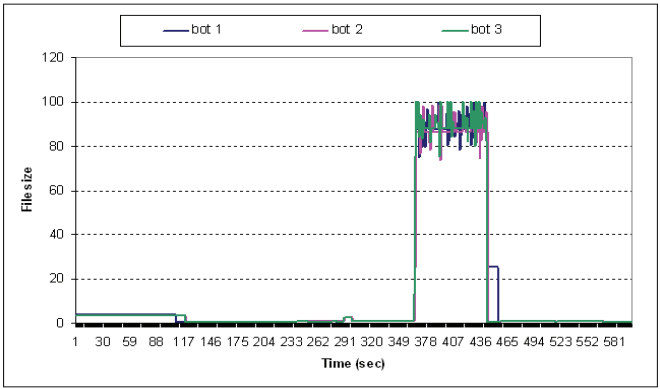

Fig. 5. Attack behaviour with UDP flood. (100 ≡ 94050 bytes)

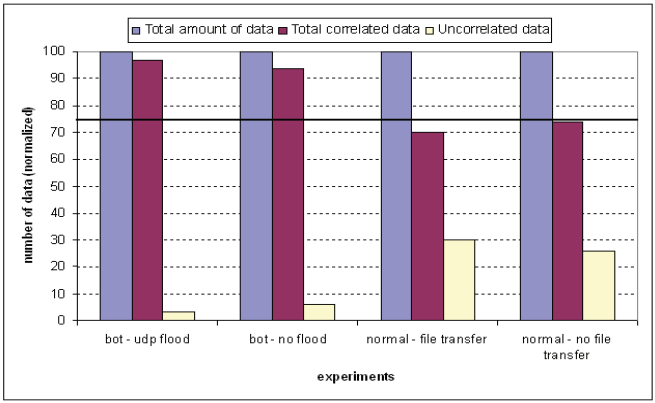

Fig. 6. Correlation between log files

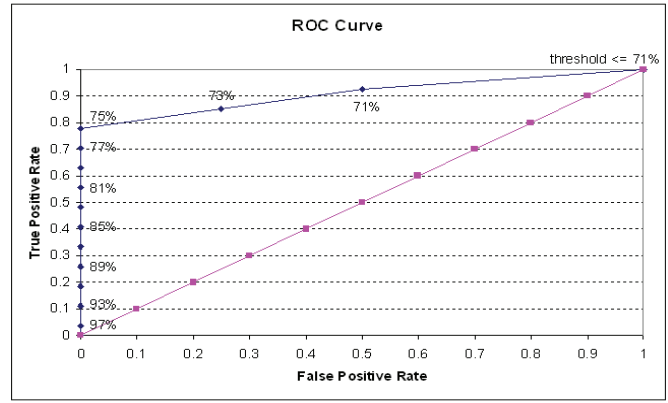

Fig. 7. The ROC curve - false positive rate vs. true positive rate. The percentages represent the threshold used.

To test how good our algorithm is in detecting botnets, we use a Receiver Operating Characteristic *(ROC)* analysis as shown in Figure 7. The x-axis represents a cumulative false positive rates while the y-axis represents the cumulative true positive rates.

We set our threshold as a percentage of log file size. As we vary the threshold from 0% to 100%, we notice that our correlation algorithm detects abnormal activity when the value of threshold is above 70% of the total amount of data and produce zero true negative. Reducing our threshold to 70% generates one false positive (i.e. normal behaviour detected as attack). Setting the threshold below 70% generates two false positive while maitaining 100% detection rate.

## V. CONCLUSION AND FUTURE WORK

Our results show that it is sometimes difficult to distinguish between normal behaviour and malicious behaviour. Therefore, we used an algorithm to detect bots based on change of behaviour by correlating events from different hosts. The correlation algorithm shows that there is a high number of correlated events in attack case generated by bots compared to normal users. Our future work will focus on detecting botnets based on not only finding correlation between events, but also monitoring the number of *API* function calls to detect a single bot in the host. We will use this approach to detect abnormal behaviour in Peer-to-Peer network.

ACKNOWLEDGMENT

This research is supported in part by the Automated Scheduling, Optimisation and Planning (ASAP), The University of Nottingham. The authors would like to thank Etisalat College of Engineering and Emirates Telecommunication Corporation (ETISALAT), United Arab Emirates, for providing financial support for this work.